\documentclass{bmvc2k}

\usepackage{times}
\usepackage{epsfig}
\usepackage{graphicx}
\usepackage{amsmath,amssymb} % define this before the line numbering.
\usepackage{algorithmic}
\usepackage[table]{xcolor} 
\usepackage{xspace}
\usepackage{enumerate}
\usepackage[ruled, lined, linesnumbered] {algorithm2e}
\usepackage{caption}
\usepackage{mathrsfs}
\usepackage{tabu}

\newlength\myindent
\newcommand\bindent{%
  \begingroup
  \setlength{\itemindent}{\myindent}
  \addtolength{\algorithmicindent}{\myindent}
}
\newcommand\eindent{\endgroup}

\newcommand{\braces}[1]{\left\{ #1 \right\}}

\newcolumntype{a}{>{\columncolor{Gray}}c}
\newcolumntype{b}{>{\columncolor{red}}c}

\title{Ensemble Manifold Segmentation for Model Distillation and Semi-supervised Learning}

\addauthor{Dengxin Dai}{dai@vision.ee.ethz.ch}{1}
\addauthor{Wen Li}{liwen@vision.ee.ethz.ch}{1}
\addauthor{Till Kroeger}{kroegert@vision.ee.ethz.ch}{1}
\addauthor{Luc Van Gool}{vangool@vision.ee.ethz.ch}{1}

\addinstitution{
 Computer Vision Laboratory\\
 ETH Zurich\\
 Switzerland
}

\runninghead{D. Dai, W. Li, T. Kroeger, L. Van Gool}{Ensemble Manifold Segmentation}

\def\ie{\emph{i.e}\bmvaOneDot}
\def\eg{\emph{e.g}\bmvaOneDot}

\begin{document}

\maketitle
\begin{abstract}

%manifold structures carry great useful information and thus have been exploited extensively for unsupervised and semi-supervised learning. 
%This paper presents a novel method for training neural networks with manifold structures. 
Manifold theory has been the central concept of many learning methods. However, learning modern CNNs with manifold structures has not raised due attention, mainly because of the inconvenience of imposing manifold structures onto the architecture of the CNNs.  
%manifold structures has not raised due attention, mainly due to the inconvenience of imposing manifold structures onto the architecture of the CNNs.  
%under the guidance of manifold structures has not received due attention. 
In this paper we present ManifoldNet, a novel method to  encourage learning of manifold-aware representations. Our approach segments the input manifold into a set of fragments. By assigning the corresponding segmentation id as a pseudo label to every sample, we convert the problem of preserving the local manifold structure into a point-wise classification task.
Due to its unsupervised nature, the segmentation tends to be noisy. We mitigate this by introducing ensemble manifold segmentation (EMS). EMS accounts for the manifold structure by dividing the training data into an ensemble of classification training sets that contain samples of local proximity. CNNs are trained on these ensembles under a multi-task learning framework to conform to the manifold. ManifoldNet can be trained with only the pseudo labels or together with task-specific labels.  
We evaluate ManifoldNet on two different tasks: network imitation (distillation) and semi-supervised learning. Our experiments show that the manifold structures are effectively utilized for both unsupervised and semi-supervised learning. 
\end{abstract}

%%%%%%%%% BODY TEXT
\section{Introduction}
\label{sec:intro}

\begin{figure}[!tb]
\centering
\includegraphics[width=0.8\textwidth]{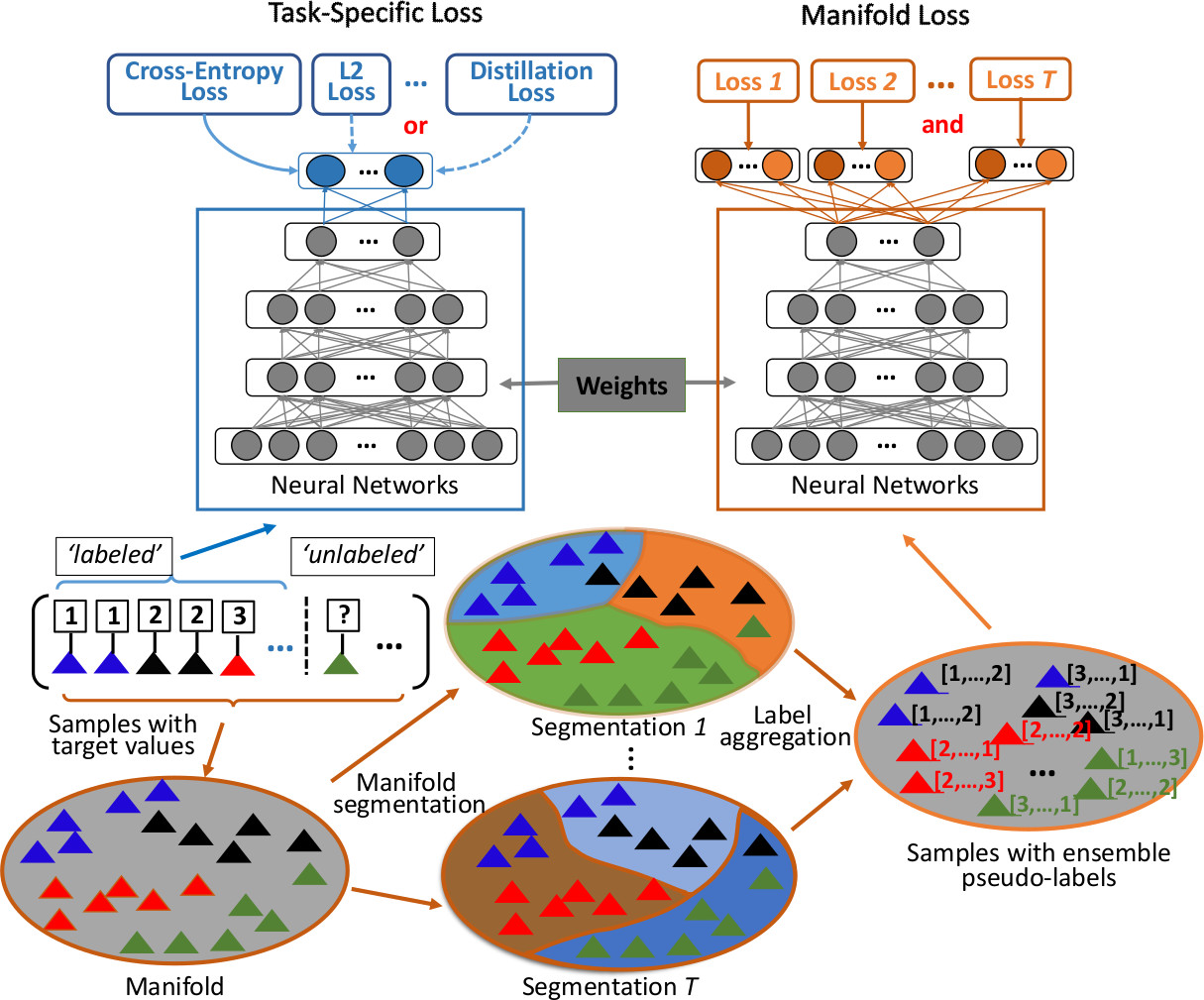} 
\vspace{2mm}
\caption{\textbf{ManifoldNet Pipeline.} The bottom panel shows ensemble manifold segmentation (EMS) for the ensemble of pseudo-labels. The top panel features the architecture of the network, which consists of two network streams shared parameters. The left stream is
guided by a task-specific loss function (\eg classification loss for semi-supervised classification and L2 loss for regression) and is trained with samples whose target values are available. The right stream is guided by an ensemble of pseudo classification tasks and trained with the unlabeled samples used for the clustering.}
\label{fig:pipeline} 
\vspace{-3mm}
\end{figure}

The frontiers of computer vision have been reshaped profoundly in the
last few years due to the ever wider deployment of convolutional neural networks
(CNNs). A cornucopia of CNNs have been developed and now define the
state of the art for many vision tasks. On the other hand, manifold learning has been a central
concept for many learning algorithms over decades, with
applications to tasks as diverse as dimensionality
reduction~\cite{lle:science:00,isomap},
hashing~\cite{manifold:hashing}, 
feature encoding~\cite{siftllc:CVPR10},
clustering~\cite{manifold:clustering,sparse:manifold:clustering},
semi-supervised learning~\cite{Belkin:semiframe:2006}, 
model imitation~\cite{dai:metric:imitation}, and
visualization~\cite{manifold:visualizing}. 
Thus, it stands to reason to devise methods to  utilize manifold structures more effectively in training CNNs. 

%under the guidance of manifold structures.
The main challenge to train a CNN with manifold structures is to incorporate
the latter's structure onto the former's architecture. Manifold structures are often
expressed by a graph or an affinity matrix of all data samples. This is inconvenient to use with many CNNs, as these are tailored for classification tasks with crisp class labels. Systems do exist for learning CNNs on graphs~\cite{bruna2013spectral,cnn:graph, learning:cnn:graph}, which accommodate affinity graphs directly as training data. However, the scalability of this strand of method is limited, since affinity graphs can potentially be large. 

%deep
%embedding~\cite{Weston:2008,manifold:cnn:14} have been developed to
%accommodate affinity matrices as training data. 

%This work aims to reconcile this gap.
In order to seamlessly couple manifold structures with the architecture
of modern CNNs, we propose to segment data manifold. It is segmented so 
that samples that are close to each other fall into the same group or 
`pseudo-class'. The corresponding pseudo-labels are fed to the CNN to 
train it for classification: grouping similar samples and separating 
dis-similar ones. This is in line with the aim of manifold learning. 
Yet, the labels obtained from the segmentation can be noisy 
due to its unsupervised nature. To mitigate this, 
we propose ensemble manifold segmentation (EMS) to create an ensemble of segmentations that are accurate individually and mutually diverse. 

%A seeds-based segmentation method was designed to fullfill these requirements. 

EMS leads to an ensemble of pseudo classification tasks, which results
in an ensemble-task architecture featuring an ensemble of loss
functions.  Figure~\ref{fig:pipeline} shows the architecture
of our method. It consists of two copies of the same network, with 
shared weight parameters. The right stream is trained with 
unlabeled data and their pseudo-labels as generated from ensemble clustering; the stream on the left is trained under `real' supervision, 
with training samples whose real target labels are available. The method is dubbed ManifoldNet. 
ManifoldNet can be trained with only the right stream or with the two 
streams jointly, depending on the nature of the tasks. 
For instance, for unsupervised learning tasks such as dimensionality reduction, hashing and unsupervised network imitation (distillation), one can use only the right stream. 
%of which the target values are generally available such as network imitation (distillation), the two streams are  trained with the same data set. 
For tasks such as semi-supervised classification, the two streams are trained with different training sets,  one labeled, one unlabeled. This flexibility greatly increases the applicability of the method.

%  (both sharing the same weights), but trained with different
% loss functions, one with task-specific loss and the other with
% ensemble manifold losses.
%that are applied to the same set or different sets of training data, depending on the task.

%, especially given the fact that many modern CNNs are tailored for classification tasks with crisp class labels. 

ManifoldNet translates manifold structures to crisp labels, which gives
representational and training advantages with modern CNNs. Apart from
being intuitive and easy to implement, ManifoldNet has additional
benefits. Compared to manifold learning methods~\cite{lle:science:00,dai:metric:imitation,isomap},  it comes naturally with an out-of-sample ability: the trained CNNs 
can be used for many tasks, in the same way as standard CNNs can, \eg 
as a feature extractor; 
Compared to deep embedding~\cite{Weston:2008,deep:kernel:reg:09, manifold:cnn:14, revisiting:semi_graph}, ManifoldNet is better scalable as it can be trained in a point-wise manner.  
ManifoldNet is a very general framework and can be easily applied to many different tasks.
As Figure~\ref{fig:pipeline} shows, much of the work needed to apply 
it to a new task lies in adopting a task-specific network 
architecture. The method is orthogonal to the  methods~\cite{doersch2015unsupervised,fl:video,DFB16} for self-learning feature representations.  
ManifoldNet is evaluated on two different tasks: network imitation and semi-supervised
classification. Experiments show that it effectively utilizes manifold structures for both unsupervised and semi-supervised learning. 

%The paper is structured as follows. Section~\ref{sec:related} presents the related work. Section~\ref{sec:method} is devoted to the approach, followed by Section~\ref{sec:tasks} for the tasks. Section~\ref{sec:conclusion} concludes the paper.

%%%%%%%%%%%%%%%%%%%%%%%%%%%%%%%%%%%%%%%%%%%%%%%%%%%%%%%%%%%%%%%%%%%%

\section{Related Work}
\label{sec:related}

%Our method is relevant to manifold learning,network learning with graph embedding, unsupervised feature learning  and ensemble learning.

\noindent
\textbf{Manifold Learning.} 
As mentioned in Section~\ref{sec:intro}, manifold learning has been
extensively studied, and numerous approaches have been developed for many different  tasks~\cite{lle:science:00,manifold:clustering,manifold:visualizing,Belkin:semiframe:2006,dai:metric:imitation}. 
%While based on different criteria, these methods share the same overall goal: reducing the complexity of the data while achieving a more understandable/manageable representation of the same information. Notable linear approaches include Principal Component Analysis, Independent Component Analysis, and Linear Discriminant Analysis. There is also a lot of of work on nonlinear manifold learning. Those methods proceed mainly in two steps: 1) quantify manifold properties (local geometry); and 2) seek a low-dimensional space so that the learned properties are mostly preserved. 
A large range of manifold properties have been
studied, such as Local Linearity~\cite{lle:science00, NPEmbedding:iccv05}, Locality ~\cite{eigenmaps:nips01}, and Local Tangent Space Analysis~\cite{ltsa}. While having a great success, these methods share a common drawback. That is, they need
to solve the Eigen system of an affinity matrix, which limits them to
problems of relatively small-scale, with notable exceptions~\cite{large:manifold}. 
%This makes large-scale manifold learning a research focus~\cite{large:manifold}.
By contrast, we incorporate the manifold structure by learning non-linear transformations via deep networks, which yield a greater expressiveness, better optimization ability and better scalability.

\noindent
\textbf{Network Learning with Graph Embedding.} 
% The learning capacity of deep neural networks couples with the
% difficulty of their training: deeper models comes with greater
% learning ability but are harder to optimize. 
%While CNNs are powerful models with a capacity to adapt to various tasks, they also require large number of training examples to prevent overfitting.  In addition to providing more training data, developing effective regularization techniques is another widely-used solution.
Graph embedding has been extensively studied as a regularization for network learning~\cite{Weston:2008,deep:kernel:reg:09, manifold:cnn:14, revisiting:semi_graph}. While based on different criteria, these methods share the same overall goal: learning feature representations for the target tasks while preserving the mutual relationships between data samples, \eg, neighbors are expected to be neighbors in the new feature space. Excellent results have been obtained for unsupervised learning~\cite{hadsell2006dimensionality} as well as semi-supervised learning~\cite{Weston:2008,revisiting:semi_graph}.
%In \cite{Weston:2008}, the authors proposed to train a neural network model jointly with an unsupervised embedding task, leading to improved  results.  In \cite{pseudo:task:08,manifold:cnn:14}, the authors regularize the networklearning by encoding the prior knowledge that a similarity matching between an image and a spatial pattern should tolerate some scaling and translation as well as slight intensity variations. The method was shown to be effective for visual recognition. This idea of learning with sensibly-designed kernels was exploited further in \cite{deep:kernel:reg:09}, with excellent theoretical support.  
Our approach has in common with these methods that it recognizes the
importance of kernel functions (manifold) for learning in deep neural
networks. At the same time, our method differs significantly from that
previous work. Those learning methods require as input the Gram matrix 
of the training samples, which is expensive in terms of storage. This 
requirement limits the scalability of these methods, as it does to all 
kernel learning approaches. Our method however, encodes the manifold 
into an ensemble of pseudo-labels, which is a much lighter representation
than the Gram matrix and is easier to feed to a CNN tailored for recognition.

% as input is the source of increased computational complexity with
% respect to the number of datapoints.
% \noindent
% \textbf{Unsupervised Feature Learning.}
% In terms of applications, our method is akin to methods which learn
% middle- or high-level image representations with unlabeled data
% \cite{stl-10,DFB16,dai:EnProDeepFets, fang2015collaborative, liao2016learning}. Clustering results of images or image patches have been leveraged to train (regularize) neural networks in the literature \cite{stl-10, liao2016learning, fang2015collaborative, yang2016joint}. Excellent results have been obtained for tasks such as image clustering and image recognition. The main difference of our method to those lies in our way of leveraging ensemble segmentation (clustering ), to aaccount for the data manifold. The drawback of individual segmentations are canceled out by working collaboratively.  \cite{DFB16} generate surrogate classes by augmenting each patch with its transformed versions and trains CNNs on top of these surrogate classes. The difference to \cite{DFB16} is that our pseudo classes are composed of a cluster of different images rather than perturbated versions of a single instance, which makes the networks more robust to intra-class variance. \cite{dai:EnProDeepFets} learns features with sampled
% image prototypes as well, but the task is not formulated as a neural
% network learning task, which leads to a inferior performance level compared to ours.

\noindent
\textbf{Ensemble Learning and Clustering.} 
Our method learns the representation from an ensemble of pseudo-labels, 
thus building itself on Ensemble Clustering (EC). Many excellent approaches have been developed for EC~\cite{strehl2002cluster, dai:ensemble:eccv12, spectral:ensemble:clustering}. EC builds a committee of base
learners and to find better solutions by maximizing their agreement. Our ensemble manifold segmentation method shares similarity with EC. Other publications close to ours
are~\cite{strehl2002cluster,ensemble:iccv11,dai:ensemble:eccv12},
where images are classified to sub-sampled categories to accumulate
distance measures in an ensemble manner. While showing some similarity,
none of the approaches learn the metrics nor the features with deep neural
networks.

% \subsection{Unsupervised Feature Learning} 
% learning feature representation by k-means~\cite{coates2012learning}
% discriminative manifold learning, 
% deep embedding,   

% bag of words supervised cnn \cite{bow:cnn}, see the three ways to
% sample positives and negatives

% \subsection{Distillation}
% %\subsection{hashing} 

% pairwise similarity, multiwise similarity, and discrete hashing via
% classification ()

% semantic separability 
% work which take graph as input: 
% Convolutional Neural Networks on Graphs
% with Fast Localized Spectral Filtering 

%%%%%%%%%%%%%%%%%%%%%%%%%%%%%%%%%%%%%%%%%%%%%%%%%%%%%%%%%%%%%%%%%%%%%%

\section{Approach} 
\label{sec:method}

In this section we first describe our ensemble manifold segmentation and then we explain how to train neural networks with the generated
pseudo-labels.

\noindent
\subsection{Ensemble Manifold Segmentation}
\label{sec:seg} 

We use an ensemble of pseudo classification tasks to incorporate prior
knowledge about manifold structures into the training of a CNN. These tasks 
need to 1) be automatically constructible from unlabeled data, and 
2)  assign two semantically 
similar images to the same pseudo label, in the spirit of manifold learning.

Given an unlabeled dataset of $N$ images, $\mathcal{D}=\{x_1, x_2, ...,
x_N\}$, we would like to generate an ensemble of pseudo-labels for
them. The number of pseudo classes is fixed in advance and is denoted by $Z$.  
Let's denote by $T$ the number of segmentation trials of the ensemble, the pseudo-labels then are $\dot{\mathcal{Y}}=\{\dot{\mathcal{Y}}_1,
\dot{\mathcal{Y}}_2, ..., \dot{\mathcal{Y}}_T\}$ and $\dot{\mathcal{Y}}_{t \in [1,T]}=\{ \dot{y}_1^t, \dot{y}_2^t, ..., \dot{y}_N^t \}$,
where $\dot{y} \in \{1, ..., Z\}$. 

As desired in manifold learning, the pseudo classes (clusters) need to be \emph{intra-pure} 
and \emph{inter-distinctive}, i.e. samples that are in the same group should
preferably belong to the same semantic class, and samples in different groups 
to different semantic classes. A straightforward solution is to directly apply a clustering algorithm such as $k$-means to the data samples. However, the assumption of isotropic distributions of samples hardly holds for high-dimensional image features. Inspired by previous works~\cite{dai:ensemble:eccv12,dai:EnPro:iccv13}, we propose a seeds-based segmentation method. Seed images that are easily separable are sampled first, and a multi-class discriminative segmentation model is then trained on the `easy' samples to quantize the whole manifold.

%Due to the unsupervised nature, the generation of this pseudo classes will be noisy. We mitigate this by using the idea of Ensemble Learning~\cite{strehl2002cluster, dai:ensemble:eccv12}, where a committee of diverse base  learners are used to compensate the drawbacks of individual learners.    

%In particular, we adopt $k$-means algorithm with random initialization as the base learner (clustering method) for our EC. 

% $k$-means algorithm algorithm is run $T$ times over the data and the clustering results are recorded as the pseudo labels. 

%

\subsubsection{Seed Image Selection}
\label{sec:seeds}

%we run the $k$-means algorithm with random initialization over all data and record the clustering results as the pseudo labels: $\dot{\mathcal{Y}}_{t \in [1,T]}=\{ \dot{y}_1^t, \dot{y}_2^t, ..., \dot{y}_N^t \}$, where $\dot{y}_n^t$ is the cluster ID of the $n$th sample $x_n$.  The set of pseudo labels $\dot{\mathcal{Y}}=\{\dot{\mathcal{Y}}_1,
%\dot{\mathcal{Y}}_2, ..., \dot{\mathcal{Y}}_T\}$ are obtained by running the $k$-means $T$ times. Other clustering algorithms such as spectral clustering can be used  in place of $k$-means as well. This will be discussed in the experiments.  
In each trial $\forall t \in [1,...,T]$, we first identify a 
small set of seed images for the segmentation. The goal is to hit as 
many different object classes as possible with these seeds, and to hit the classes safely -- avoiding cases where seeds are
positioned near class boundaries.
Due to the unsupervised nature, we keep the method simple and general. 
We now present the seed selection heuristic.
The algorithm has its basis in the assumptions that seed images 
are surrounded densely by their neighbors.  
On the one hand, the algorithm tries to spread out the seed images, 
\ie at a relatively large distance from each other (as with manifold 
learning). On the other hand, ensemble learning theory suggests that 
members of the ensemble should be highly varied in order to form a 
powerful committee. This implies that at each round, the seed
images should spread out with respect to a standard distance
measure, and across different rounds, the sampled seed images
should be diverse. 

We leverage $k$-means clustering algorithm for this. In each trial, $k$-means is randomly initialized to cluster all samples in $\mathcal{D}$ to $Z$ clusters. After convergence, we position seeds to the cluster centres, leading to a small pseudo datasets: $\mathcal{S}_t=\{(x_{c_1},1), (x_{c_2},2), ..., (x_{c_Z}, Z)\}$. 
Although it is possible to train a discriminative
segmentation (classification) method with one training sample per
pseudo-class, \ie the sampled seed images, it is beneficial to enrich
the data from a single seed image to a prototype set of multiple images 
for the purpose of learning better intra-class
invariance~\cite{singh2012unsupervised,dai:ensemble:eccv12,DFB16}. To this end, the $K$
nearest neighbors of each seed image are introduced into the
corresponding pseudo class, \ie the included neighbors are given the
same pseudo-label as the seed image. This data enrichment is in line
with manifold learning: neighboring samples are expected to
share the same labels. The $Z$ image prototypes with their pseudo-labels
are then used to train a discriminative classifier to segment the
whole manifold. The seed growing step provides richer
training data for the discriminative learning, while avoiding to
considerably degrade data quality. The flaws of each such training
set are compensated for by those of the other training sets, as 
they are diverse.

%The procedure is similar to curriculum learning~\cite{Curriculum_learning}, where learning starts with easier tasks, to then tackle harder ones. 

\subsubsection{Ensemble Manifold Segmentation} 

We now explore the utility of the seed sets acquired in 
Section~\ref{sec:seeds} to gather information that can be useful for 
the manifold segmentation.  In particular, we train a multi-class 
discriminative classifier for each trial $t$: $\phi_t(\cdot): x 
\rightarrow \dot{y}$,  $\dot{y} \in \braces{1, \ldots, Z}$ for each 
seed set $\mathcal{S}_t$. The full data $\mathcal{D}$ is then classified 
by $\phi_t$ to give a class label for every sample: $\dot{y}_i = 
\phi_t(x_i)$.  The resulting class labels of all samples in $\mathcal{D}$ 
in trial $t$ is $\dot{\mathcal{Y}}_t$ and the pseudo labels over all $T$ 
trials are denoted as $\dot{\mathcal{Y}}=\{ \dot{\mathcal{Y}}_1, ..., 
\dot{\mathcal{Y}}_t\}$.  

The complete algorithm for ensemble manifold segmentation is summarized
in Algorithm~\ref{alg:1}, where $\mathcal{N}[x_i, k]$ returns the
$k^{th}$ closest neighbor for image $x_i$. We
employ logistic regression for the classifier $\phi_t$ as it is 
efficient and also generalizes well when trained only on a few
training samples, which is an important characteristic as we kept the
size of the individual seed sets relatively small. The resulting ensemble
pseudo-labels $\dot{\mathcal{Y}}$ are then used to train neural
networks for various vision tasks. It is noteworthy to mention that 
the dataset $\mathcal{D}$ used so far does not require any labels, meaning
that no real target values (\eg class labels, regression values) are 
used. Therefore, this dataset can be the same as or different from 
the dataset that comes with target values for an associated supervised 
learning task (left stream of the network in Figure~\ref{fig:pipeline}).  
The learning procedure will be discussed in Section \ref{sec:cnntraining}. 

%Most of the bottom panel of Figure~\ref{fig:pipeline} has now been explained.

\begin{algorithm} [tb]
% \label{alg_EnProx}
\caption{Ensemble Manifold Segmentation}
\label{alg:1}
\begin{algorithmic}
\STATE \textbf{Input}: the dataset $\mathcal{D}$.
\STATE \textbf{Output}: pseudo-labels $\dot{\mathcal{Y}}=\{\dot{\mathcal{Y}}_1, \dot{\mathcal{Y}}_2, ..., \dot{\mathcal{Y}}_T\}$   \\
\FOR{$t=1 \to T$} 
    \STATE 0. Seed positioning: sample $\mathcal{S}_t$ by randomly-initialized $k$-means;
    \STATE 1. Seed growing: \\ 
    \bindent 
    \FOR{$z=1 \to Z$} 
    	\FOR{$k=1 \to K$} 
           \STATE $\mathcal{S}_t  \leftarrow  \mathcal{S}_t \cup \{ (\mathcal{N}[x_{c_l},k],z)  \}$
        \ENDFOR
    \ENDFOR
    \eindent 
    %$ \vect{s}_i^t = \vect{s} \oplus \bold{ind}, \vect{c} = \vect{c} \oplus \vect{i}$;\
    \STATE 2. Train classifier $\phi_t(\cdot) \in \braces{1\ldots Z}$ on $\mathcal{S}_t$.
    \STATE 3. Segmenting entire $\mathcal{D}$ by $\phi_t(\cdot)$ for $\dot{\mathcal{Y}}_t$: 
    \bindent \STATE $\dot{\mathcal{Y}}_t = \{ \dot{y}_{1,t},  \dot{y}_{2,t}..., \dot{y}_{N,t} \}$ and $\dot{y} \in \{1,...,Z\}$ .    \eindent
\ENDFOR
\end{algorithmic}
\end{algorithm}

\subsection{Learning CNNs with Ensemble Pseudo-Labels}
\label{sec:cnntraining}

Our goal in this section is to train a CNN to make a prediction 
$f: \mathcal{X} \rightarrow \mathcal{Y}$. In the classification case, 
$\mathcal{Y} = \{1, ..., C\}$ with $C$ the number of classes, and 
$l(.)$ is usually the cross-entropy loss function. For a regression 
task, $\mathcal{Y}=\mathbb{R}$, and $l(.)$ can be the $L_2$ or $L_1$ 
loss. Standard supervised training proceeds as follows: given a training 
set with $M$ samples $\mathcal{D}_{s}=\{(x_1, y_1), ..., (x_M, y_M)\}$, 
it minimizes the loss  
\begin{equation}
\label{eq:target}
L_s(\theta_{s}) = \sum_{i=1}^M l(y_i, f(x_i, \theta_{s}))  + \lambda_s \|\theta_{s}\|.  
\end{equation}
where $\lambda_s$ is a scaling factor for the regularization and $\theta_{s}$ is the parameters of the model $f$. In our scenario, this optimization corresponds to learning the left stream of the network in Figure~\ref{fig:pipeline}.
%The optimization is done by a stochastic gradient descent algorithm.  

Similarly, training a neural network with the generated pseudo-labels 
$\dot{\mathcal{Y}}$ extracted from the manifold in Section~\ref{sec:seg} can be formulated as minimizing the cost 
\begin{equation}
\label{eq:manifold} 
L_m(\theta_{m}) = \sum_{t=1}^T \sum_{i=1}^N l(\dot{y}_{i}^t, \dot{f}_t(x_i, \theta_{mt}))  + \lambda_m \|\theta_{mt}\|. 
\end{equation}
where $\lambda_m$ is a regularization parameter and $\theta_{mt}$ is the parameters of the model $\dot{f}_t$ for manifold learning. Note that all networks share all the parameters except for the last layer.  
This optimization corresponds to learning the right stream of the 
network in Figure~\ref{fig:pipeline}. The CNN learned with this cost can 
be considered as a deep counterpart of the conventional manifold learning 
techniques, such as LLE and ISOMAP, and it can be applied to the same 
set of applications, such as dimensionality reduction and hashing.   

Our pseudo tasks can be considered to encode prior knowledge into a 
manifold, that can be incorporated into the supervised learning by adding 
extra cost functions in a multi-task learning framework, as shown in 
Figure~\ref{fig:pipeline}. These two learning tasks then jointly train 
the network. This architecture is especially suitable for semi-supervised 
learning, where labeled data are used to train the left network in Figure~\ref{fig:pipeline}, following Equation~\ref{eq:target}, 
and unlabeled data for the right network, following Equation~\ref{eq:manifold}. This architecture is also useful for tasks where 
all samples have target values, such as network distillation, as the knowledge represented by the manifold is 
complementary to the knowledge from these target values. 
Learning the network in this case equals to minimizing 
\begin{equation}
\label{eq:cost} 
L(\theta) =  L_s(\theta_s) + \lambda L_m(\theta_m)
\end{equation}
where $\lambda$ is to balance the two cost terms. 

The architectures and parameter settings of 
the networks are explained in Section~\ref{sec:tasks} in the context of 
specific tasks. 

%and dimensionality reduction using an auto-encoder, 
% As can be seen from Figure\ref{fig:pipeline}, the parameters of ManifoldNet 
% consists of three parts: the shared parameters $\theta_s$ by the two streams, 
% the parameters $\theta_{t}$ for the task-specific loss, and the parameters 
% $\theta_{m}$ for the ensemble manifold losses. The main interest of the 
% learning is to optimize the shared parameters $\theta_s$, by the supervision 
% of the pseudo-labels alone, or together with the task-specific loss.  For 
% the sake of generality, we describe the method as the two streams are 
% trained jointly.

%The parameters are learned by minimizing three terms: 1) task-specific loss, 
% 2) ensemble manifold losses, and the regularization term:
%\begin{equation}
%  \mathcal{L}  = \mathcal{L}_{t}(\mathcal{D}_{su}, \mathcal{Y}) + \lambda %\sum_{t=1}^T % \mathcal{L}_{m}(\mathcal{D}_{m}, \dot{\mathcal{Y}}_t) + \mathcal{R}%%%(\theta_s, 
% \theta_t, \theta_m)
%\end{equation}
%where $\lambda$ is a scaling factor to balance the two lost functions;
%the regularization $\mathcal{R}$ corresponds to a simple weight decay
%of $0.0005$. We train the network to minimize this loss function using
%stochastic gradient descent. 
%The complete training process of the
%network is summarized in Table 1~\ref{table1}. 

\section{Learning Tasks} 
\label{sec:tasks}

%We describe the learning of CNNs for a variety of tasks, based on  the pseudo-labels generated in Section\ref{sec:seg}. These tasks include  dimensionality reduction (DR), network distillation~\cite{hinton2015distilling},  and semi-supervised classification. For the first two tasks,  $\mathcal{D}_{s}$ and $\mathcal{D}$ are the same dataset, henceforth both will be denoted by $\mathcal{D}$. For semi-supervised learning, they are  different and $\mathcal{D}_{su}$ can be expected to be much smaller  than $\mathcal{D}$ as manual annotations are costly. 

In this work, we evaluate our method in the context of two
different tasks: model imitation (unsupervised network distillation), and semi-supervised classification. The same parameter
choices are made for all the tasks. Concerning the network training,
$\lambda_s$ and $\lambda_m$ correspond to a simple weight decay of $0.0005$;
$\lambda$ is set to a value such that the losses from the two
streams are approximately the same. The parameters of the segmentation
method are set as follows: $Z=30$, $T=90$ and $K=9$. 

The datasets considered are standard datasets for image classification
and retrieval: 
STL-10~\cite{stl-10},  CIFAR-100~\cite{cifar}, Scene-15~\cite{lazebnik:cvpr06}, Indoor~\cite{Indoor}, and SUN397~\cite{SUN397}. The details about
their usage will be described along with the corresponding tasks.

\subsection{Model Imitation} 
Obtaining models of computational efficiency at test time is crucial,
especially for applications where running time and storage space are
at a premium. Among others, model imitation (compression or
distillation)~\cite{model:compression,dai:metric:imitation,hinton2015distilling}
is one of the most popular approaches towards addressing this problem.
The main idea is to use a fast and compact model to imitate the
function learned by a better-performing, but slower and larger model.
The learned function is expressed usually by its responses over data
samples~\cite{model:compression,hinton2015distilling} or by a manifold
over the data samples~\cite{dai:metric:imitation}.  In this work, we
adopt a manifold as the expression.

\bgroup
\def\arraystretch{1}%  1 is the default, change whatever you need
\begin{table*}[!tb]
  \centering \small \setlength{\tabcolsep}{.18em} 
  \caption{The accuracy (\%) of image retrieval at a recall of $1$. The goal is to fine-tune the pre-trained AlexNet to imitate the pre-trained VGGNet and ResNet as well as possible.} \vspace{2mm}
      \rowcolors{8}{gray!25}{white}
      \begin{tabu}{c|[2pt]cccc|[2pt]cccc}
       \rowcolor{gray!50}
    &  & \multicolumn{2}{|c|}{\textbf{AlexNet $\Rightarrow$ VGGNet}} &  &   & \multicolumn{2}{|c|}{ \textbf{AlexNet $\Rightarrow$ ResNet} }  &   \\   
    & AlexNet & \cite{hinton2015distilling} &  Ours & VGGNet & AlexNet & \cite{hinton2015distilling} &  Ours  & ResNet   \\   \hline  
    STL-10    & 42.59 &  46.23  & \textbf{51.56}  & 52.12  &  42.59  & 47.43  & \textbf{51.39}  & 55.80 \\ 
    CIFAR-100 & 5.40  & 5.37   & \textbf{6.07}   &  5.70 & 5.40  & 11.45 & \textbf{6.07}  & 7.75 \\ 
%    Indoor-67 &       &     &     &   &   & &   &  \\ 
 %   SUN-397   &       &     &     &  &   &   &   &  \\ 
   \end{tabu}
    \label{tab:mi:retrieval} 
    \vspace{-2mm}
\end{table*}
 \egroup

   %  0.4259    0.5212    0.5580    0.4495    0.5139    0.2816
  
\subsubsection{Shallow Network Imitates Deep Network} 
Usually, deeper networks have a better
learning ability but are memory-hungry and slow. We investigate
the problem of using shallow networks to imitate
deeper networks. We take AlexNet as our shallow network, and VGGNet
and ResNet as our deep networks. Given a collection of images,
the pairwise distances among them are computed with features from the deep
networks (L2 is used as well); these images are then segmented into
ensemble pseudo classes; ManifoldNet is trained according to
Eq.~\ref{eq:manifold}, \ie the right stream in
Fig.~\ref{fig:pipeline}. The network architecture is instantiated
with an AlexNet with its classification layer replaced by an ensemble
of pseudo classification layers. The network is then fine-tuned
according to the losses of all the pseudo classification tasks.

We compare our method with network
distillation~\cite{hinton2015distilling}, which is akin to
our method, but imitates the output scores of the teacher (deep)
networks. It is noteworthy that our primary interest is to perform
model imitation in an `unsupervised' manner, where no labels of the
images are used.
%  We also compare our method to deep CCA~\cite{}, which learns Canonical
% Correlation Analysis between two features with neural networks.
The task is evaluated on STL-10 and CIFAR-100, for image
retrieval. The model is trained with the images from the training sets
without using their labels, and the performance is reported on the test sets. 

% For the
% other two, the training is performed on one half of the images and the
% test is on the other half. The split is done randomly.

The retrieval results are listed in Table~\ref{tab:mi:retrieval}.
The table shows that both methods improve the student (shallow)
network by imitating the `response' of the teacher (deep) network over
a collection of unlabeled images. This implies that even
unlabeled data samples can be used to transfer knowledge. 
The table also shows that
our method outperforms network
distillation~\cite{hinton2015distilling} for this task. 
We acknowledge that the network distillation problem~\cite{hinton2015distilling} is different from our model imitation considered here, as the latter requires knowledge transfer across datasets as well. Therefore, a comparison to~\cite{hinton2015distilling} on this task needs to be treated with caution. Nevertheless, \cite{hinton2015distilling} has been widely used for this type of knowledge transfer  even across different data modalities, \eg from images to depth~\cite{supervision:transfer}, and from video
to audio~\cite{video2sound}.  
The advantage of ManifoldNet over network distillation for image retrieval is due to the fact that the learning goal of ManifoldNet is in line with distance preservation. Thus, the features it produces are more suitable for similarity measurement. 

%ManifoldNet is also trained with both streams for this task, where the left stream is for the distillation ~\cite{hinton2015distilling}, while the right stream is for the ensemble pseudo classification. We find this joint training can further improve the results. This implies that the two techniques exploit complementary information, and can be combined.  

% four or five figures as the number of training data changes, the
% classification performance increase.
  
\subsection{Semi-supervised Classification}
Recent years have witnessed considerable progress in image recognition
due to the advance of deep
learning~\cite{deepnet:nips12,vgg16,resnet}. These methods, however,
heavily rely on the quantity of labeled training data. The
explosion of visual data, combined with the high cost of human
annotation, starts shifting focus towards learning with less supervision. One typical
example is semi-supervised classification (SSC)~\cite{Weston:2008,dai:EnPro:iccv13, kingma2014semi, springenberg2015unsupervised, ladder:network, distributional:smoothing, improved:GAN, hoffer2016semi}. which aims to learn better classifiers with the help of unlabeled
data. ManifoldNet is here evaluated for SSC. 

Before diving into the experiments, we need to group previous methods into two groups. The first group consists of all manifold-based methods: to `regularize' the learners with  manifold structures of unlabeled samples. The methods include Laplacian SVM (LSVM)~\cite{LapSVM}, Harmonic-Function 
(HF)~\cite{Zhu:Harmonic:03}, Ensemble Projection
(EP)~\cite{dai:EnPro:iccv13}, and Deep Semi-supervised Embedding
(DeepSSE)~\cite{Weston:2008}. A comparison to these methods is an apple-to-apple comparison with the aim to show which methods can utilize manifold structures the best. 
The second group is generative model based approaches~\cite{kingma2014semi, springenberg2015unsupervised, ladder:network, distributional:smoothing, improved:GAN}, which build on the recent success of generative (adversarial) models. They `regularize' the classifier by adding an auxiliary task to the original classification task, such as denoising in \cite{ladder:network} and fake-real image classification in \cite{improved:GAN}. While these methods define the state-of-the-art, they are currently limited to datasets of small images such as CIFAR-10~\cite{cifar} and SVHN~\cite{svhn} due to the computational complexity. We compare with them on CIFAR-10 and more importantly study the complementarity between our method and those approaches. 

\noindent
\textbf{Manifold-based Methods}. We compare with manifold-based methods on four classification datasets:  STL-10, CIFAR-100, Indoor-67~\cite{Indoor} and SUN-397~\cite{SUN397}.  
The AlexNet and VGGNet pre-trained on ImageNet are used for feature extraction and network
fine-tuning. Six competing methods are considered: Logistic Regression
(LR), a baseline of Fine-tuned Networks
(FTuning), LSVM, HF, EP, and DeepSSE. For LR, LSVM, HF, and EP, the features are directly
extracted with the pre-trained AlexNet and VGGNet. For FTuning, the
two pre-trained networks are fine-tuned with the annotated
data. DeepSSE and ManifoldNet fine-tune the two networks with both
labeled and unlabeled data: DeepSSE adds an auxiliary embedding task,
while ManifoldNet adds an ensemble of pseudo classification
tasks. DeepSSE and ManifoldNet are both implemented with the
architecture of AlexNet and VGGNet.

 \bgroup
\def\arraystretch{1}% 1 is the default, change whatever you need
\begin{table*}[!tb]
  \centering \small \setlength{\tabcolsep}{.15em} 
  \caption{Accuracy of image classification on four datasets. $30$ images per class are used as the labeled data, with the rest of the training images as unlabeled data (see text for details). Missing values are due to the high computational complexity.}
  \vspace{2mm}
  \rowcolors{8}{gray!25}{white}
      \begin{tabu}{c|[1.5pt]ccccccc|[1.5pt]ccccccc}
        \rowcolor{gray!50}
        &  \multicolumn{7}{c|[1.5pt]}{\textbf{AlexNet}}   & \multicolumn{7}{c}{ \textbf{VGGNet} }    \\
 & LR & \cite{LapSVM} & \cite{Zhu:Harmonic:03} & \cite{dai:EnPro:iccv13} & FTuning & \cite{Weston:2008}  & Ours  & 
LR & \cite{LapSVM} & \cite{Zhu:Harmonic:03} & \cite{dai:EnPro:iccv13} & FTuning & \cite{Weston:2008}  & Ours  \\ \hline 
  %     SVHN      &53.7&53.2&43.4&54.8 &54.0&58.3& \textbf{67.1} & 55.2 &56.0&48.6& 56.8   & 55.3 & 62.0 & \textbf{69.8}\\
       STL-10  &85.5&85.6&80.4&86.0 &85.6&86.9& \textbf{87.2} & 91.7 &87.4&86.3& 92.1   & 91.6 & 91.6 & \textbf{92.4}\\
        CIFAR-100 &34.0&35.1&34.2&35.8 &34.1&37.0& \textbf{40.2} & 36.2 &37.2&34.9& 39.2   & 44.1 & 44.1 & \textbf{52.9}\\
        Indoor-67 &50.4&51.2&46.9&52.2 &50.0&53.3& \textbf{57.8} & 62.5 &63.4&62.1& 64.1   & 63.1 & 64.8 & \textbf{67.5}\\
        SUN-397   &43.5& ---&---&43.2  &44.1&44.7& \textbf{45.8} & 50.3 & ---&--- & 50.4   & 50.0 & 51.0 & \textbf{52.3} \\
     \end{tabu}
    \label{tab:ssl:classification}  
\end{table*}
\egroup

%For STL-10, CIFAR-100, and SVHN, the models are trained on theirtraining sets, with $30$ images per class as labeled data and the rest as unlabeled data. The classification accuracy is reported on the testing sets. 

% \begin{figure*} [!th]
%   \centering
%    $ \begin{array}{cccc}
% \hspace{-1mm}
% \includegraphics[width=0.33\linewidth]{./T-crop.pdf}& \hspace{-2mm}  
% \includegraphics[width=0.33\linewidth]{./Z-crop.pdf}& \hspace{-2mm} 
% \includegraphics[width=0.33\linewidth]{./K-crop.pdf}& \hspace{-2mm} \\
% \text{(a) the size of the ensemble} & \text{ (b) The number of seeds used} & \text{(c) The number of neighbors included}  \\
% \end{array}$
% \caption{Model imitation evaluated on STL-10 for image retrieval, where AlexNet is used to imitate ResNet.}
%   \label{fig:parameter}  
% \end{figure*}

For STL-10 and CIFAR-100, all images of the training sets are used for the training, out of which $30$ images per class are used as labeled data.  
For both Indoor-67 and SUN-397, $80\%$ of the images per class are used as the
training data, out of which $30$ images per class are taken as labeled
data and the rest as unlabeled data. The performance is reported
on the other $20\%$ images. 
The classification results are listed in
Table~\ref{tab:ssl:classification}.  The table shows that our method
outperforms the competing methods of the first group significantly and consistently. The
superior performance to LR, LSVM, HF, and EP is mainly due to the fact
that ManifoldNet learns task-specific features deeply while these
methods train classification models with pre-trained, general features
directly. This is also supported by the fact that DeepSSE outperforms
these methods as well. 

The table also shows that simply
training (fine-tuning) a network on the new data does not improve the
performance when the amount of labeled data is limited. $30$ images
per class are not very little, but are still insufficient. This is exactly the scenario where DeepSSE and our
method comes to the rescue with a sensibly-designed regularization term from
the manifold structures of a large collection of unlabeled data. The
table shows that these data-driven auxiliary tasks do help in
regularizing the neural networks. 
Compared to DeepSSE, ManifoldNet turns the `embedding' task into a classification
task. Instead of penalizing individual pairs, it maintains an
explicit model of the manifold structures with the pseudo
classes. The representation provides richer contextual insights of the
neighborhood structure. Also, the quadratic growth of the number of
pairs hinders the learning efficiency of DeepSSE. 

%2) the representation of the pseudo classes are discrete labels, which are
%representationally simple and make the learning more robust than
%regressing continuous values directly. Distance values of image pairs
%are not always consistent, so regressing them directly hinders the
%convergence of the training. Recent works have also found that
%discrete class labels are advantageous over continuous
%values~\cite{flow:static,pixelrnn} for other network learning
%tasks. 

\noindent
\textbf{Generative Model Based Methods.} 
We compare with the three generative model based methods:  Ladder~\cite{ladder:network}, CatCAN~\cite{springenberg2015unsupervised} and iGAN~\cite{improved:GAN}. Our network architecture is the same as the discriminative part of iGAN \cite{improved:GAN}. Since all these methods learns the network from scratch, we adapt our method to learn without fine-tuning. It is an iterative process. Initially, the manifold is constructed with the SIFT\_llc feature~\cite{wang2010locality}. ManifoldNet is trained using its two streams. After the first round of training, a new manifold will be generated using the just trained features for another round of training. The iteration continues three times as running more times does not provide further improvement.   

The results on CIFAR-10 are reported in Table~\ref{tab:ssl:classification2}, where $400$ images per class are used as the labeled data. As the table shows, our method achieves comparable (slightly worse) results to the state-of-the-art methods without using any pre-trained model. The iterative training procedure provides an improvement of $10\%$ and the procedure can be used to `jointly' learn classifiers and manifold structures. Since our method is complementary by nature to these generative model based methods. We further look into combining the two. For this, we start with the trained iGAN model, augment it our pseudo classification tasks, and then train them jointly. The combination improves the accuracy of iGAN from $81.37\%$ to $83.38\%$.  

 \bgroup
\def\arraystretch{1.2}% 1 is the default, change whatever you need
\begin{table}[!tb]
  \centering \small \setlength{\tabcolsep}{.20em} 
  \caption{Accuracy of image classification on CIFAR-10. } \vspace{2mm}
  \rowcolors{8}{gray!25}{white}
      \begin{tabu}{cccc|c}
        \rowcolor{gray!50}
        Ladder~\cite{ladder:network} & CatCAN~\cite{springenberg2015unsupervised} & iGAN\cite{improved:GAN} & Ours &  iGAN+Ours  \\ \hline 
        79.60 & 80.42 & \underline{81.37} & 79.7 & \textbf{83.38} \\ \hline 
     \end{tabu}
    \label{tab:ssl:classification2}  \vspace{-2mm}
\end{table}
\egroup

\section{Conclusion}
\label{sec:conclusion}
In this work, we have proposed a novel method ManifoldNet to learn
neural networks with manifold structures from unlabeled data.
ManifoldNet translates the graph-based manifold structures into an
ensemble of pseudo classification labels, significantly easing the
process of coupling the structure of manifolds and deep neural
networks. ManifoldNet can be trained with the generated pseudo
classification labels only or with manual annotations jointly.  This
gives it the flexibility to be applied to various of applications. This
work evaluated the method on two tasks: model imitation, and semi-supervised classification. Experiments show that ManifoldNet effectively utilizes manifold structures for both unsupervised and semi-supervised learning. 

{\small
\bibliographystyle{ieee}
\bibliography{egbib}
}

\end{document}